\let\TeXyear\year
\documentclass{ieeeaccess}

\let\year\TeXyear

\usepackage{comment}
\usepackage{color}
\usepackage{tikz}
\usetikzlibrary{shapes,arrows,automata, positioning}

\definecolor{accessblue}{cmyk}{1, 0.3, 0, 0.2}
\definecolor{greycolor}{cmyk}{0,0,0,.8}
\NewSpotColorSpace{PANTONE}
\AddSpotColor{PANTONE} {PANTONE3015C} {PANTONE\SpotSpace 3015\SpotSpace C} {1 0.3 0 0.2}
\SetPageColorSpace{PANTONE}
\usepackage{dblfloatfix}

\usepackage{algorithm,algpseudocode}
\usepackage{listings,multicol}  
\usepackage[many]{tcolorbox}

\tcbuselibrary{listings}
\usepackage[space]{cite}
\usepackage{svg}
\usepackage{multirow}
\usepackage[font=normalsize]{caption}
\usepackage{float}
\usepackage{tikz}
\usepackage[left=2cm,right=2cm,bottom=2cm,top=2cm]{geometry}
\usepackage{listings}
\usepackage{euscript}
\usepackage{amssymb}
\usepackage{placeins}
\usepackage{bytefield}
\usepackage{amsmath}
\newcommand{\colorbitbox}[3]{%
\rlap{\bitbox{#2}{\color{#1}\rule{\width}{\height}}}%
\bitbox{#2}{#3}}

\definecolor{lightcyan}{rgb}{0.84,1,1}
\definecolor{lightgreen}{rgb}{0.64,1,0.71}
\definecolor{darkgreen}{rgb}{0,0.6,0}
\definecolor{lightred}{rgb}{1,0.7,0.71}
\definecolor{asparago}{rgb}{0.48,0.62,0.36}
\definecolor{lightgreen}{rgb}{0.64,0.64,0.71}
\definecolor{lightblue}{rgb}{0,0.0,0.75}
\definecolor{amber}{rgb}{0.9, 0.75, 0.0}
\definecolor{darkamber}{rgb}{0.8, 0.65, 0.0}

\definecolor{javared}{rgb}{0.6,0,0} % for strings
\definecolor{javagreen}{rgb}{0.25,0.5,0.35} % comments
\definecolor{javapurple}{rgb}{0.5,0,0.35} % keywords
\definecolor{javadocblue}{rgb}{0.25,0.35,0.75} % javadoc
\definecolor{solarized@base03}{HTML}{002B36}
\definecolor{solarized@base02}{HTML}{073642}
\definecolor{solarized@base01}{HTML}{586e75}
\definecolor{solarized@base00}{HTML}{657b83}
\definecolor{solarized@base0}{HTML}{839496}
\definecolor{solarized@base1}{HTML}{93a1a1}
\definecolor{solarized@base2}{HTML}{EEE8D5}
\definecolor{solarized@base3}{HTML}{FDF6E3}
\definecolor{solarized@yellow}{HTML}{B58900}
\definecolor{solarized@orange}{HTML}{CB4B16}
\definecolor{solarized@red}{HTML}{DC322F}
\definecolor{solarized@magenta}{HTML}{D33682}
\definecolor{solarized@violet}{HTML}{6C71C4}
\definecolor{solarized@blue}{HTML}{268BD2}
\definecolor{solarized@cyan}{HTML}{2AA198}
\definecolor{solarized@green}{HTML}{859900}
\usepackage{graphicx} 
\usepackage{hyperref}
\graphicspath{ {res/} }

\PassOptionsToPackage{normalem}{ulem}
\usepackage{ulem}

\providecolor{hl}{rgb}{0,0,0.9}
\providecolor{hln}{rgb}{0,0,0.9}
\providecolor{deleted}{rgb}{1,0,0}

\newcommand{\mytilde}{\raise.17ex\hbox{$\scriptstyle\mathtt{\sim}$}}

\usepackage{fix-cm}    

\makeatletter
\newcommand\HUGE{\@setfontsize\Huge{20}{25}}
\makeatother

\def\BibTeX{{\rm B\kern-.05em{\sc i\kern-.025em b}\kern-.08em
    T\kern-.1667em\lower.7ex\hbox{E}\kern-.125emX}}

\tikzstyle{block} = [draw, fill=green!20, rectangle, 
   minimum height=10em, minimum width=13.2em]
\tikzstyle{sum} = [draw, fill=blue!20, circle]
\tikzstyle{input} = [coordinate]
\tikzstyle{output} = [coordinate]
\tikzstyle{pinstyle} = [pin edge={to-,thin,black}]

\begin{document}
\history{Date of publication xxxx 00, 0000, date of current version xxxx 00, 0000.}
\doi{10.1109/ACCESS.2023.0322000}

\title{Compressed Real Numbers for AI: a case-study using a RISC-V CPU}
\author{
\uppercase{Federico Rossi}\authorrefmark{1},
\uppercase{Marco Cococcioni}\authorrefmark{1}, \IEEEmembership{Senior Member, IEEE},
\uppercase{Roger~Ferrer~Ibáñez}\authorrefmark{2}, 
\uppercase{Jes\'us~Labarta}\authorrefmark{2}, 
\uppercase{Filippo~Mantovani}\authorrefmark{2}, 
\uppercase{Marc~Casas}\authorrefmark{2}, 
\uppercase{Emanuele Ruffaldi}\authorrefmark{3}, \IEEEmembership{Senior Member, IEEE} and
\uppercase{Sergio Saponara}\authorrefmark{1}, \IEEEmembership{Senior Member, IEEE},
}

\address[1]{Department of Information Engineering, University of Pisa, Pisa, Italy (e-mail: federico.rossi@ing.unipi.it, {marco.cococcioni, sergio.saponara}@unipi.it}
\address[2]{Barcelona Supercomputing Center, 08034 Barcelona, Spain, (e-mail: \{roger.ferrer, jesus.labarta, filippo.mantovani, marc.casas\}@bsc.es)}
\address[3]{MMI spa, (e-mail: emanuele.ruffaldi@mmimicro.com)}
\tfootnote{ Work partially supported by the following H2020 projects: EPI SGA2 (grant agreement 101036168), EuPilot (grant agreement 101034126) and TextaRossa (grant agreement 956831).}

\markboth
{Author \headeretal: Preparation of Papers for IEEE TRANSACTIONS and JOURNALS}
{Author \headeretal: Preparation of Papers for IEEE TRANSACTIONS and JOURNALS}

\corresp{Corresponding author: Federico Rossi (e-mail: federico.rossi@ ing.unipi.it).}

\begin{abstract}
As recently demonstrated, Deep Neural Networks (DNN), usually trained using
single precision IEEE 754 floating point numbers (\emph{binary32}), can also work
using lower precision.
Therefore, 16-bit and 8-bit compressed format have attracted considerable attention. In this paper, we focused on two families of formats that have already achieved interesting results in compressing binary32 numbers in machine learning applications, without sensible degradation of the accuracy:  \emph{bfloat} and \emph{posit}. Even if 16-bit and 8-bit bfloat/posit are routinely used for reducing the storage of the weights/biases of trained DNNs, the inference still often happens on the 32-bit FPU of the CPU (especially if GPUs are not available). In this paper we propose a way to decompress a tensor of bfloat/posits just before computations, i.e., after the compressed operands have been loaded within the vector registers of a vector capable CPU, in order to save bandwidth usage and increase cache efficiency. Finally, we show the architectural parameters and considerations under which this solution is advantageous with respect to the uncompressed one.
\end{abstract}


\titlepgskip=-21pt

\maketitle

\section{Introduction}
\label{sec:introduction}

In the last years the RISC-V open-source architecture has been put under the spotlight \cite{riscvisa,waterman2011risc,asanovic2014instruction}. Thanks to its open nature and absence of royalties, it is making its way among the big microprocessor companies like Intel, AMD, and Arm. Moreover, several hardware and software companies are contributing to the initiative, other than public and private universities around the globe.

Thanks to its modularity, the RISC-V architecture offers multiple extensions for the instruction set architecture (ISA). Furthermore, it provides the possibility to customise and expand the base ISA. Among the interesting ISA extensions developed for this new architecture, the vector extension is one of the most advanced and promising for high-performance computing (HPC) and machine learning. In particular, this instruction set extension 
%tries to fill the gap with 
complements
pre-existing architectures like SSE/AVX (Streaming SIMD Extension/Advanced Vector eXtension) by Intel, and SVE (Scalable Vector Extension) by Arm.

All these extensions also aim to overcome the absence of GPUs/TPUs (Graphics/Tensor Processing Units) in several hardware ecosystem, while promoting a coherent programming paradigm across the architectures. Moreover, thanks to the vector length agnostic approach taken by Arm and RISC-V, the portability of applications can skyrocket in the same architecture environment, without the need to adapt each program for different configurations of the same architecture.

In combination with these new advancements in machine learning and HPC applications, several new optimisations in information representation have been proposed by both industry and academia. In particular, the focus has shifted to reducing computation and space complexity in handling floating-point computations, which are pervasive in the machine learning world. Among the others, some deserve important credits: Intel with Flexpoint \cite{koster2017flexpoint,popescu2018flexpoint}, Google with bfloat16 \cite{burgess2019bfloat}, IBM with DLFloat16 \cite{IBM_DL_Float16_ARITH19}, Facebook AI with logarithmic numbers \cite{johnson2018rethink}, and Tesla with Configurable Floats \cite{tesla_configurable_float}. 

In this paper, we mainly focus on Google bfloat and on another, very promising, number system: the posit{\texttrademark} format. Posit numbers have been proposed by J. L. Gustafson \cite{gustafson2017beating} and proven to match the accuracy obtained when using 32-bit floating point numbers (i.e., the so called \emph{binary32}, according to the IEEE 754/2019 standard) with that achieved by 16-bit posits \cite{coco2019icecs,coco2018exploiting,deeppositron,positnn,Lu_et_al_2020,coco_et_al_ieeespm_2020,coco2023smallreals}. In the meantime, several researchers have pushed towards hardware implementations of this new format, showing how it can be interesting in terms of power consumption and on-chip area utilisation \cite{dinechin2019eval, coco2020lppu}.

Our goal in this work is to combine both the two emerging approaches (a vectorized RISC-V processor and the arithmetic compression), where a smaller format is used as a compressed storage format to be used in a big length (up to 16384-bit) vectorized CPU. The combination between a highly compressed format (down to 16-bits, so up to a factor 2 with respect to a standard binary32 representation for weights and activations) and a large vector register environment can be particularly interesting when the architecture allows to fit entire matrices or vectors inside a single vector register. Moving 16-bit operands instead of 32-bit operands improves the memory/CPU data transfer bandwidth, has a positive impact on all the caches, and allows to put more operands within the vectorized environment at the same time.

Furthermore, in the paper, we push towards testing the capabilities of this approach in a hardware platform, exploiting an advanced model of a real RISC-V vector processor. We use the Vehave emulator and MUlti-level Simulation Approach (MUSA) simulator to evaluate the impact of our approach.

%\IEEEPARstart{R}

\subsection{Contributions of this work}
In this paper we envision the use of compressed numerical formats to store and transfer data, to be used in the RISC-V vector environment; the work is
aimed at extending what have been done in \cite{coco_et_al_nca}. More precisely, in the present extension we address two topics not addressed before: the use of alternative numerical formats for DNN \emph{data compression} (weights and activations) on common operations when training DNNs (i.e., the matrix-matrix multiplication and convolution), for providing feedbacks to RISC-V Register Transfer Level (RTL) designers. More precisely, we want to asses under which conditions the decompression ("widening") of the weights/activations can be delayed at the latest, i.e., only once the data has been loaded within the CPU, and by using vectorized instructions for decompression/compression.
Similarly to \cite{coco_et_al_nca},
even in this work we assume to not have GPUs/TPUs, a situation which is not uncommon. The presence of GPUs/TPUs will be addressed in a future study.

\subsection{Organisation of the paper}

In Section \ref{sec:riscvvect} we present the RISC-V vector extension and we describe on the emulation and simulation ecosystem, including an overview of the tools used in the subsequent sections.

In Section III we present the bfloat and the posit formats, the two compressed format for DNN weights/biases considered in this work.

In Section \ref{sec:in_register_compression} we present the proposed approach, namely, the in-register compression and decompression method to deal with 16-bit real number formats.

In Section \ref{sec:results} we
present the obtained results on a GEMM kernel with different matrix sizes and vector register lengths, evaluating the impact of the instruction set architecture on the overall latency.

Finally, we draw conclusions in section \ref{sec:concl}.

\section{The RISC-V processor and the ``V" extension}\label{sec:riscvvect}
The key aspect of the RISC-V Instruction Set Architecture (ISA) is its modularity \cite{riscvisa}. Indeed, we can build whichever combination of RISC-V ISA subsets to match application requirements. 

In this paper we will focus on the vector extension, namely, ``V" extension. This extension aims to provide a Single Instruction Multiple Data (SIMD) unit to the RISC-V processor architecture.

The current lack of hardware implementations of the RISC-V vector extensions can be mitigated by using a set of tools that runs on top of a basic RISC-V processor, either physical or emulated through virtualisation software such as QEMU (Quick EMUlator, \url{https://www.qemu.org/}). In this paper we will use the RISC-V 64-bit (\texttt{RV64GC}) processor Sifive U74-MC. Hereafter we will describe the set of tools used to emulate and analyze the behaviour of vectorize programs on a RISC-V vector capable processor (using the $1.0$ revision of the vector extension). 

\subsection{The Vehave emulator}

Vehave is a user-space emulator for the V-extension of the RISC-V ISA that runs on RISC-V Linux. 
It allows a functional verification of a program that uses V-extension instructions or a code generator, such as a compiler, that emits V-extension instructions. 
The goal is to help porting applications that run in a well-known environment, such as Linux. 
%
%The user porting an application can perform experiments with instructions included in the V-extension.
Only a compiler or code generator that can emit V-extension instructions is needed.

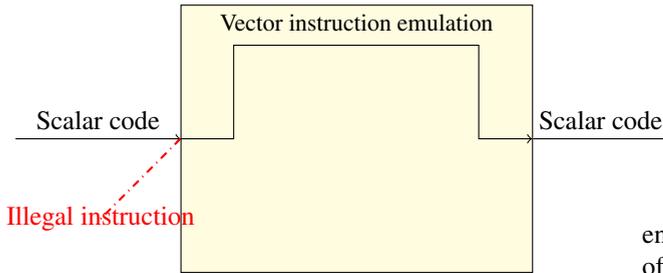
\begin{figure}[htbp]
  \centering
 
\begin{tikzpicture}[auto, node distance=2cm]
    % We start by placing the blocks
    \node [input, name=input1] {};
 
    \node [block, right of=input1, right = 0.5em,fill=yellow!15] (vehave) {};
    
    \node[below] at (vehave.north) {\small Vector instruction emulation};
    
    \node [output, name=output1, right of=vehave, right =6em] {};
    
    % Once the nodes are placed, connecting them is easy. 
     \draw [draw,->] (input1) -- node {Scalar code} (vehave);
        \draw [draw,->] (vehave) -- node {Scalar code} (output1);
    
    \draw[draw,dash dot,thick,red] (vehave.west) -- +(-3em,-3em) node[] {Illegal instruction};
    
     \draw [draw,->] (vehave.west) -- +(2em,0em) -- +(2em,3.5em) -- +(11.2em,3.5em) -- +(11.2em,0em) -- +(13.2em,0em) node[] {};
     %\draw [draw,->] (input2) -- node {$p2$} (abs2);

\end{tikzpicture}

  \caption{Graphical representation of the Vehave emulation scope (yellow).}
  \label{figVehaveTrap}
\end{figure}

Vehave emulates instructions by intercepting the illegal instruction exception that a CPU emits when it encounters an unknown/invalid instruction. 
Once an illegal instruction is found, Vehave decodes it and if it is a valid V-extension instruction it emulates it, otherwise an error is propagated back.
The program resumes once the emulation of the vector instruction is complete.
Figure~\ref{figVehaveTrap} provides a simplified graphical representation of the emulation dynamic of Vehave.

Vehave relies on the LLVM libraries of the compiler which already supports the V-extension for the process of decoding the instructions.
The output of Vehave is collected in a {\tt .trace} file which stores in plain text extensive details about each vector instruction emulated and some quantitative figure of the scalar code executed before each vector instruction.

%%%%%%%%%%%%%%%%%%%%%%%%%%%%%%%%%%%%%%%%%%%%%%%%%%%%%%%%%%%%%%%%%%%%%%%%%%%%%%%%
% OLD Vehave intro
%%%%%%%%%%%%%%%%%%%%%%%%%%%%%%%%%%%%%%%%%%%%%%%%%%%%%%%%%%%%%%%%%%%%%%%%%%%%%%%%
% Vehave is a vector extension simulator that runs on top of a 64-bit RISC-V
% processor that does not have vector capabilities. Basically, Vehave is
% instructed to trap all the \textit{illegal instruction} exception risen from
% the underlying platform. Thanks to this, vector instruction are actually
% implemented inside Vehave and, most importantly, a series of detailed
% information about instruction execution is output in a trace file.  This file
% includes information about type of instruction (opcode), source and
% destination registers and other useful information.

\subsection{The Paraver Trace Analyzer}

The {\tt .trace} file generated by Vehave are parsed and converted to {\tt .prv} format which can be visualised with Paraver~\cite{pillet1995paraver}.
Paraver is a visualisation tool originally developed to allow a qualitative global perception of the behavior of complex parallel HPC applications previously run acquiring traces on a cluster of compute nodes.
Even if the interposition library to gather the information stored in the trace is different in an HPC cluster and in Vehave, the navigation features of Paraver are flexible enough to allow studying almost any timestamped series of data converted to the {\tt .prv} format~\cite{mantovani2018performance}.
%
% The traces elaborated from the Vehave simulator can then be visualised using the Paraver tool. 

Paraver offers a graphical interface to analyze several aspects of the vector program execution.
The two main visualization configurations are timelines and histograms (or tables).
Timelines allow visualizing (with different colors) the data encoded in the trace such as: the value of the program counter, the content of the registers used by each vector instruction, and the addresses of memory accesses.
Histograms are used to quantitatively inspect the content of the traces and allow to easily collect the number and type of vector instructions as well as the register utilization.

%Starting from the Vehave traces we can analyze the overall vector register utilization during execution as well as  memory access pattern and number and type of instructions. 

\subsection{The MUSA Trace Enricher}

While Vehave has full visibility on vector instructions and their scope in the code, it does not allow to collect performance information.
The vehave emulator produces a trace of dynamic instructions without any performance estimation. The MUSA trace enricher incorporates a timing model capability in the Vehave emulation infrastructure that uses the initial Paraver trace to drive a micro-architecture simulation.
From this simulation, it enriches the trace by adding precise information in terms of instruction latency, memory hierarchy level where accesses are served, or commit cycle time.
The MUSA timing model keeps all information present in the first Paraver trace unaltered and adds additional timing and performance data.
On top of the data stored in the trace, MUSA considers different architectural abstractions to simulate fundamental aspects like memory hierarchy, instruction dependencies, or out-of-order pipeline.
%
%Several information may be missing from the Vehave traces after the simulation. Indeed, Vehave does not include all the architectural aspects of a RISC-V vector unit. The MUSA tool was designed to address this issue. 

Using textual configuration files we can instruct MUSA to simulate several architecture configurations such as instruction latency for all the vector instructions, number and size of cache levels, memory access latency and size, number of re-order buffer entries (ROB size) and other similar information about the underlying architecture.
%
%We can feed Vehave traces to this tool to enrich them with additional information such as  actual CPU execution cycles, cache miss and hit rations and average CPU cycle per instruction.

The output of the MUSA trace enricher can be analysed using Paraver as well, thoroughly visualising all the architecture behaviour during the execution of a vector program coupled with timing information.
This way we are able to estimate  both 
{\em i)} the performance of the vector implementation of our algorithms, and
{\em ii)} the effect of architectural changes on our code.

\section{Compressed formats}
In this section we review the two compressed formats investigated in this study: bfloat and posit (while bfloat is deeply investigated in this work, posit is only theoretically evaluated at the end, comparing its complexity to the bfloat format). These two formats were demonstrated to be able to replace binary32 format in deep learning tasks with very little degradation in accuracy, while reducing the memory footprint by a factor 2 \cite{coco2023smallreals}.

\subsection{The bfloat format}
The bfloat16 format \cite{burgess2019bfloat} is a 16-bit fixed-size format. It has 3 fields as shown hereafter:

\begin{figure}[t]
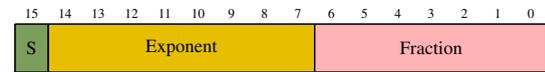

	\centering    
    \begin{bytefield}[bitwidth=1.25em]{16}
       \bitheader[endianness=big]{0-15}\\
       \small
       \colorbitbox{asparago}{1}{{\scriptsize{S}}}&
       \colorbitbox{amber}{8}{\scriptsize{Exponent}} &
       \colorbitbox{lightred}{7}{\scriptsize{Fraction}} 
    \end{bytefield}
    \caption{Illustration of a bfloat16 number.}
	\label{fig:bfloat16ill}
\end{figure}

\begin{figure}[t]
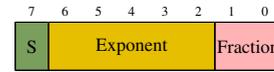

	\centering    
    \begin{bytefield}[bitwidth=1.25em]{8}
       \bitheader[endianness=big]{0-7}\\
       \small
       \colorbitbox{asparago}{1}{{\scriptsize{S}}}&
       \colorbitbox{amber}{5}{\scriptsize{Exponent}} &
       \colorbitbox{lightred}{2}{\scriptsize{Fraction}} 
    \end{bytefield}
    \caption{Illustration of a bfloat8 number.}
	\label{fig:bfloat8ill}
\end{figure}

\begin{itemize}
    \item Sign field: 1-bit
    \item Exponent field: 8-bit (same as the binary32 format)
    \item Fraction: 7-bit
\end{itemize}

The format relaxes several IEEE binary32 standard features to allow low-cost hardware implementations. In particular: (i) subnormals are clamped to 0; (ii) only one rounding mode is supported; (iii) NaN are not propagated; (iv) no IEEE exception flags.

The format can particularly fit to neural networks as an easy replacement for IEEE binary32 numbers. The main advantages of this  format are: (i) drop-in replacement for IEEE binary32 with comparable neural network accuracy; (ii) smaller footprint, hence increased throughput with regards to IEEE binary32 format; (iii) conversion between IEEE binary32 number and bfloat16 can be performed with a right-shift or a right zero padding, respectively.\\  

The bfloat family also includes a 8-bit format, namely bfloat8. The format is a fixed-size one, with 3 fields as shown hereafter:
\begin{itemize}
    \item Sign field: 1-bit
    \item Exponent field: 5-bit 
    \item Fraction field: 2-bit
\end{itemize}

As we can see from the format, the bfloat8 cannot be obtained by a simple bit-shift of the 16-bit counterpart. Instead, it can be obtained with a right shift of a binary16 (a.k.a floating point 16, fp16) number by 8 positions. This means that, in an architecture with hardware support for binary16 numbers, we can exploit this property to easily convert between binary16 and bfloat8 numbers.

\subsection{The posit format}

\begin{figure}[b]
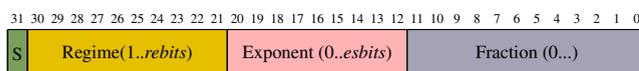

	\centering    
    \begin{bytefield}[bitwidth=0.75em]{32}
       \bitheader[endianness=big]{0-31}\\
       \small
       \colorbitbox{asparago}{1}{{\scriptsize{S}}}&
       \colorbitbox{amber}{10}{\scriptsize{Regime(1..$rebits$)}} &
       \colorbitbox{lightred}{9}{\scriptsize{Exponent (0..$esbits$)}} &
       \colorbitbox{lightgreen}{12}{\scriptsize{Fraction (0...)}} 
    \end{bytefield}
    \caption{Illustration of a posit$\left<32,9\right>$  data type. Both the exponent and the fraction field can be absent, for specific configurations having a regime field particularly lengthy.}
	\label{fig:32bitPositIllustration}
\end{figure}

The posit format \cite{gustafson2017beating,coco2020sensors,coco2019ftanh,coco_fast_elu_smartcomp} is a  fixed length format, where the total length and exponent length can be configured. It can have a maximum of 4 fields as in Figure \ref{fig:32bitPositIllustration}. The fields in a posit number are:
\begin{itemize}
    \item Sign field: 1 bit;
    \item Regime field: variable length, composed by a string of bits equal to 1 or 0 ended, respectively by a 0 or 1 bit;
    \item Exponent field: at most \textit{esbits} bits (it can even be absent);
    \item Fraction field: variable length fractional part of the significand (it can even be absent too).
\end{itemize}

Given a $posit\left<nbits,esbits\right>$, represented in 2's complement signed integer $X$ and let $e$ and $f$ be exponent and fraction values, the real number $r$ represented by $X$ encoding is:
\begin{equation*}\label{eqn:rtp}
r= 
\begin{cases}
0\text{, if $X$}=0 \\ 
\text{NaR} \text{, if $X$}=-2^{(nbits-1)} \quad \quad \;\; \text{  (Not a Real)} \\ 
sign(X)\cdot useed^k \cdot 2^e \cdot (1 + f) \text{, otherwise.}
\end{cases}
\end{equation*}

The value $useed$ is  $useed = 2^{2^{esbits}}$ and the number $k$ is the value of the regime. The value $k$ is strictly related to the regime length $l$ and bitstring ($b$ is the bit that composes the string of identical bits, e.g. in $00001$ $b=0$). If $b=0$ the $k$ is negative, otherwise the $k$ is positive:
    \begin{equation*}\label{eqn:kpu}
    k= 
    \begin{cases}
    -l\text{, if $b$}=0 \\ 
   l-1 \text{, otherwise.}
    \end{cases}
    \end{equation*}
    
In \cite{coco_et_al_jrtip_2020} we brought this kind of implementation into the vectorization field,  implementing several posit operations such as ELU and Tanh exploiting the already existent vector integer operations in the RISC-V vector environment.

\section{In-register compression and decompression}\label{sec:in_register_compression}

The core idea of the approach we propose in this paper is the decompression of compressed numbers to be performed inside the RISC-V vector registers exploiting vector instructions. In this way we aim to reduce the memory transfer to the vector registers; indeed, instead of loading $N$ elements of 32-bit, we will load the same $N$ elements but compressed on 16-bit.

\begin{figure}
    \centering
    \includegraphics[width=\linewidth]{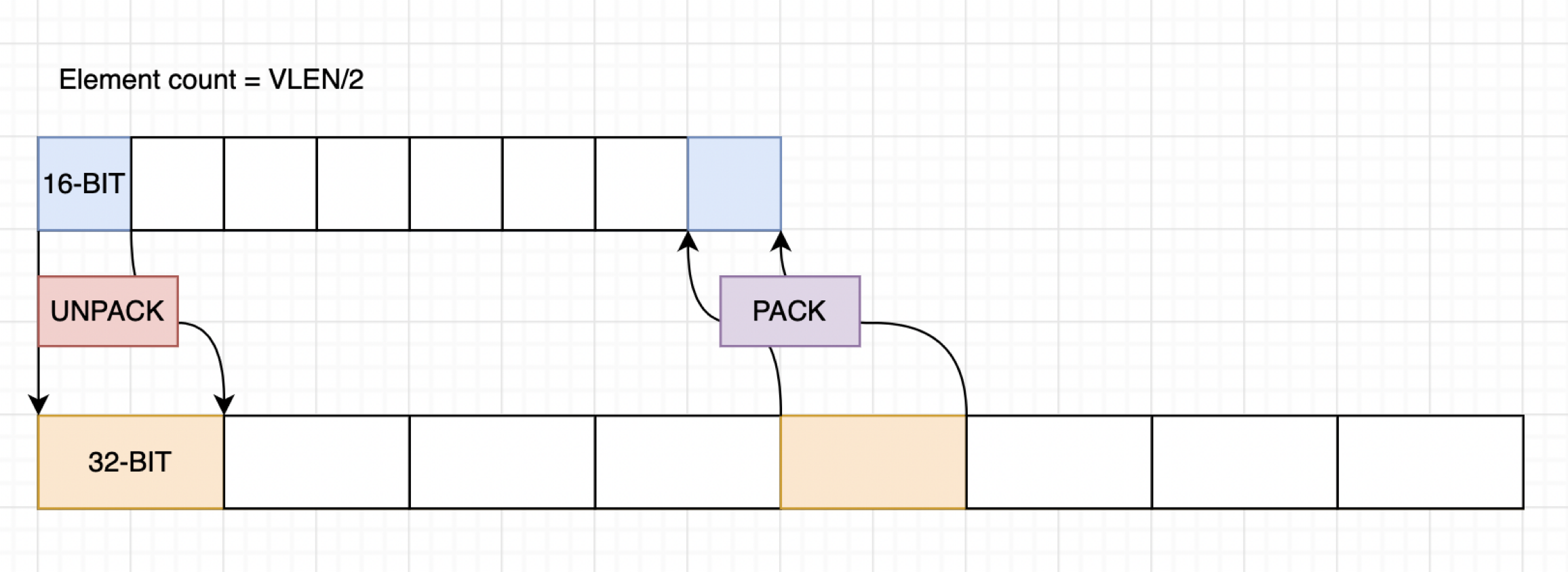}
    \caption{Process of unpacking and packing a vector of a generic compressed format directly inside a VPU register.}
    \label{fig:pack_unpack}
\end{figure}

In particular, as shown in Figure \ref{fig:pack_unpack}, when we need to decompress (or unpack) a vector filled with bfloat16 elements we start reasoning about the vector register length.

Given that we need to expand a 16-bit format inside a 32-bit one, when filling the vector register we fill just half of the vector register. Then we can expand each vector element using the RISC-V vector instructions.

Same approach is considered when we need to compress the binary32 format back to the original 16-bit one.

While this approach can be generalized to any compressed 16-bit format, it may perform very differently when applied to different low precision format.

In particular, the unpack and pack phases may involve several different instructions to perform the necessary bit manipulations for the type conversions.

While a Posit$\left<16,2\right>$ can have a very little drop in accuracy when converted from binary32, the conversion involves the use of several bit manipulations to be completed. This means that the un/pack phases can potentially take a lot of computing cycles. On the other hand, if we could have specific hardware instruction for vectorized posit to float (and viceversa) conversions, we could sensibly reduce the latency of said operations.

There are other formats, such as bfloat16, that are designed to allow seamless conversion with the binary32 format.

In particular the two phases involve following instructions:
\begin{itemize}
    \item binary32 $\rightarrow$ bfloat16: logic right shift of 16 digits
    \item bfloat16 $\rightarrow$ binary32: left shift of 16 digits with 0 padding on the least significant bits
\end{itemize}

Since the two steps have an inherent cost for the conversion we need to take into account the computation effort spent on and decompression and compression.

\begin{figure}
    \centering
    \includegraphics[width=\linewidth]{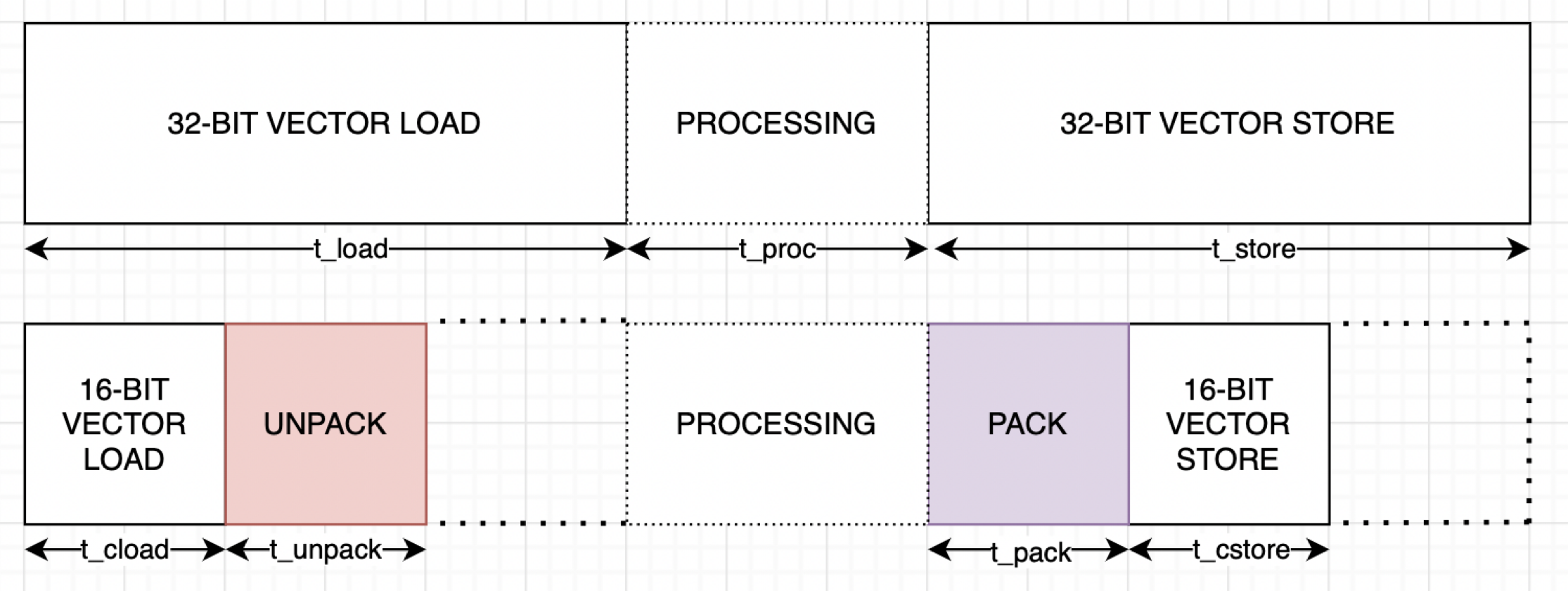}
    \caption{Timeline of a generic linear algebra algorithm that employs in-register decompression and compression.}
    \label{fig:comp_timeline}
\end{figure}

In Figure \ref{fig:comp_timeline} we show our approach in detail. Suppose we need to process some data with a given algorithm, whose processing time will be $t_{proc}$ on a given architecture. If we imagine to load the $N$ elements to be processed using a 32-bit representation we will need (on average) $t_{load}$. 

On the other hand we can imagine that, on average the load for the same number of elements, represented on 16-bit, will need $t_{cload}$, where $t_{cload} \sim t_{load} / 2$.

In the second case we will also need to pack and unpack the compressed format so we will add a $t_{pack}$ and $t_{unpack}$ to the timeline.

At this point we can compare the two timelines and write the following requirements:

\begin{equation}
    t_{cload} + t_{unpack} < t_{load}
\end{equation}
\begin{equation}
    t_{pack} + t_{cstore} < t_{store}
\end{equation}

If both these requirements hold, we can obtain an advantage on computation time using this approach.

A further advantage we can theoretically see with our approach is the possibility to exploit instruction overlapping offered by an out-of-order vector processor.

\begin{figure}
    \centering
    \includegraphics[width=\linewidth]{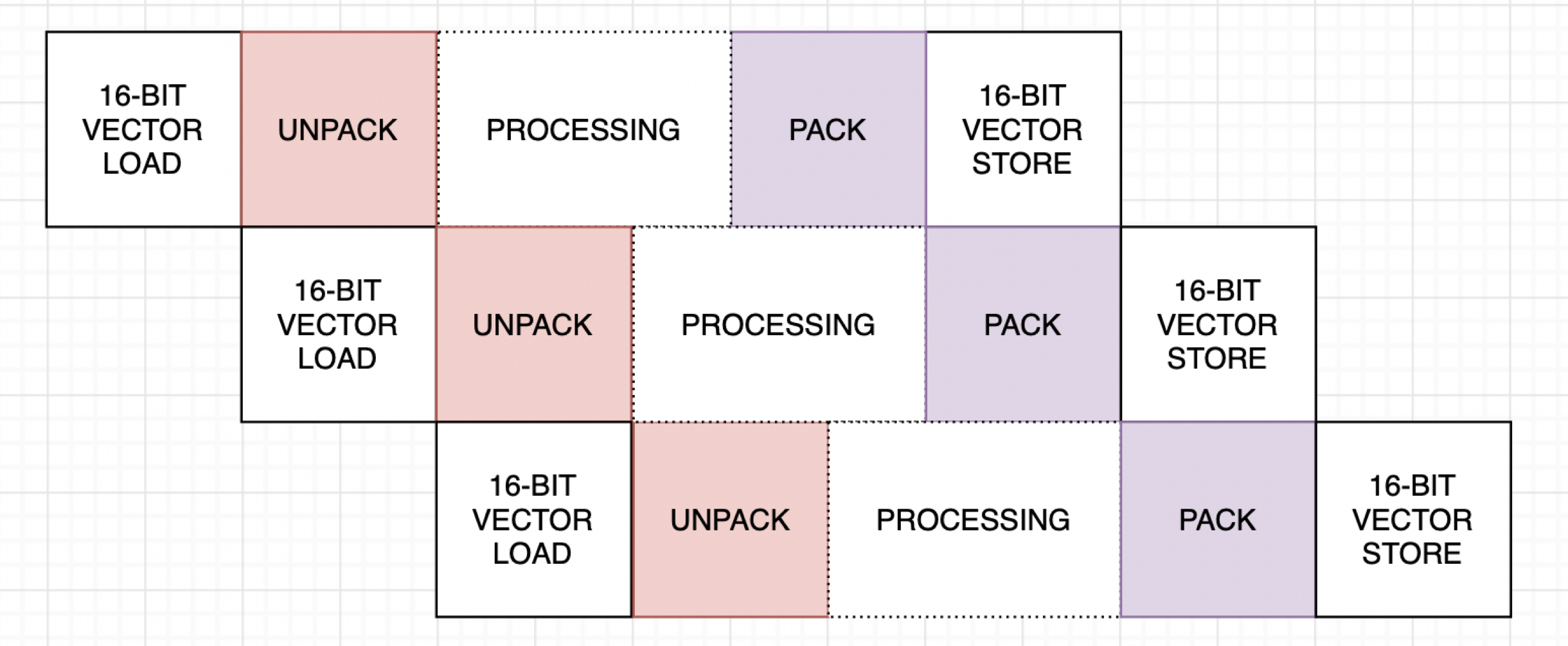}
    \caption{Possibility of interleaving the sequence of instructions seen in Figure \ref{fig:comp_timeline}.}
    \label{fig:comp_pipelined}
\end{figure}

In particular, consider that we are dealing with successive independent computations of separate pieces of data. We can imagine that a 16-bit vector load for the $(n+1)$-th piece of data can be performed while the unpacking of the $n$-th piece of data. 

As shown in Figure \ref{fig:comp_pipelined}, the same idea can be applied to the different independent parts of the overall process. While the same concept would be the same even in the 32-bit load case, note that having smaller parts of independent processing phases on data helps the interleaving of the instructions. Indeed, given that a 32-bit load takes twice as long than the 16-bit counterpart, the interleaving will start with an increased latency.

\begin{figure}
    \centering
    \includegraphics[width=\linewidth]{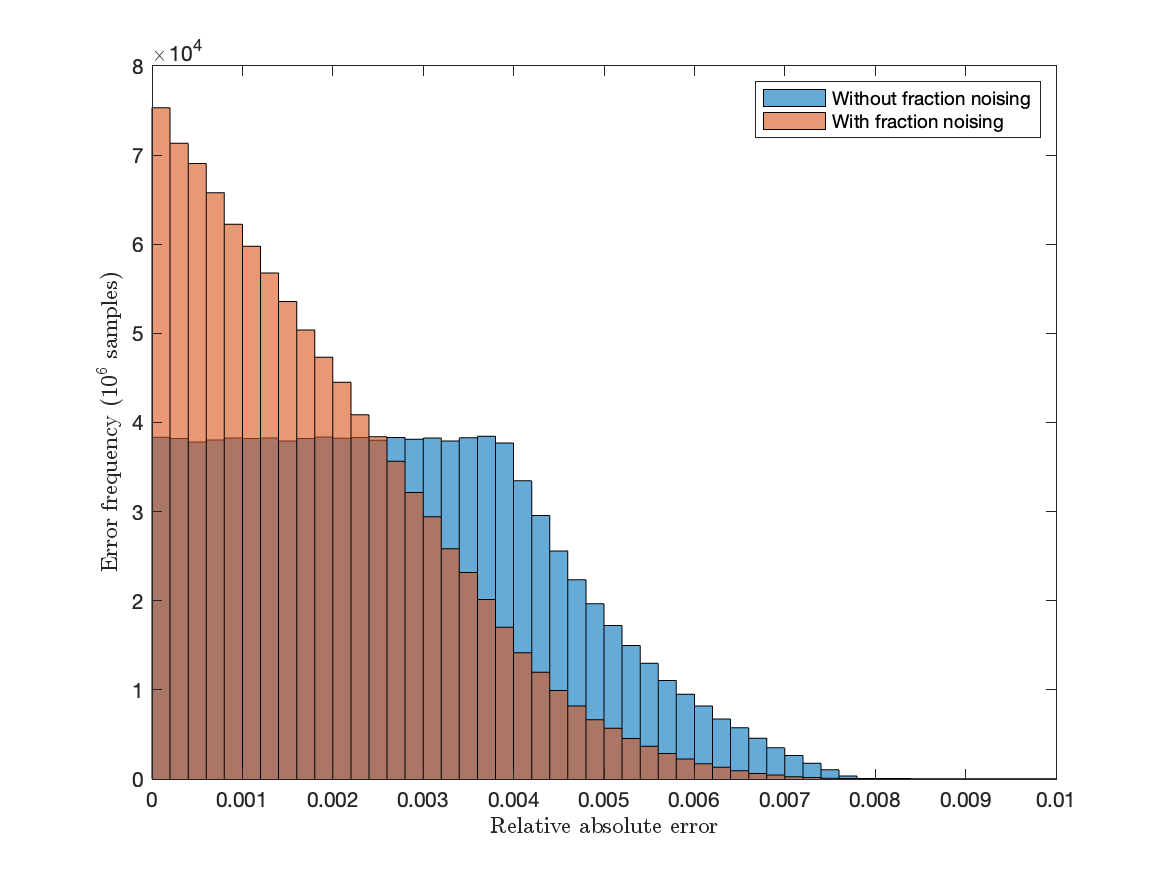}
    \caption{Relative absolute error density distribution in conversion between IEEE 32-bit floating points and 16-bit bfloats, computed on $10^6$ random samples. }
    \label{fig:fp_bf_accu}
\end{figure}

In terms of accuracy, since every operation is executed using 32-bit floating point arithmetic, the loss in precision only happens when converting from 32-bit float into the 16-bit format. Figure \ref{fig:fp_bf_accu} shows the density distribution of the absolute relative error that happens during conversion between 16-bit bfloats and 32-bit IEEE floating points. The blue curve in Figure \ref{fig:fp_bf_accu} shows the error distribution when converting from FP32 from/to BF16 ignoring the least 16 significant bits of FP32. The red curve in Figure \ref{fig:fp_bf_accu} shows the same distribution after we have applied a noise to the FP32 fraction when converting from BF16 to FP32: instead of padding 16 zeroes to the right of the BF16 we simply reproduce the 16 most significant bits into the 16 least significant bits. As reported in the two plots, this technique sensibly reduce the relative error during the conversion forth and back.

\section{Methodology and Experimental Results}\label{sec:results}

In this section we aim to demonstrate in which conditions our approach can outperform the traditional one (i.e. whether a 16-bit compressed load \& decompression outperforms the 32-bit uncompressed load).
We focus on the number of committed instructions, computing cycles and average instruction latency, provided by the Vehave simulation tool.
We tested the following algorithms, equipped with decompression/compression approach, on a matrix-matrix multiplication kernel.

Listing \ref{lst:gemm} shows an example of the proposed approach for the GEMM case. We started from a generic GEMM algorithm and modified it to directly load bunch of 16-bit compressed matrix lines and then we decompress them in-register, before computing the multiplication.

In particular, Listing \ref{lst:unpack} and \ref{lst:pack} show the actual implementation of decompression and compression techniques.

\subsection{Methodology}
In order to evaluate the goodness of our approach we compare the in-register de/compression approach to a generic kernel that does not employ compression. As a recap:
\begin{itemize}
    \item Compressed approach: images/weights are stored as bfloat16 numbers. During kernel execution, data is loaded in bunch of 16-bit elements inside the VPU registers and then it is decompressed to binary32 expanding it to occupy the remaining space inside vector registers. Computations are carried out using binary32 vector instructions already implemented by the RISC-V vector extension.
    \item Uncompressed approach: images/weights are stored as binary32 numbers. Everything is loaded uncompressed and computation is carried out on binary32 data.
\end{itemize}

In both cases we monitor the number of computing cycles spent to carry out the kernel several times.
\begin{lstlisting}[float=*t, label=lst:gemm,caption={GEMM algorithm with in-register de/compression.}]
void gemm_16(...) {
	const unsigned long int rvcntw = __builtin_epi_vsetvlmax(__epi_e32,__epi_m1);                 
	for (unsigned long int i = 0; i < M; ++i)

		for (long unsigned int j = 0; j < N; j +=  rvcntw) {

			unsigned long int all_gvl = __builtin_epi_vsetvlmax(__epi_e32,__epi_m1);
			__epi_4xf32 accum = __builtin_epi_vfmv_v_f_4xf32(0.0f,all_gvl);
			unsigned long int gvl = __builtin_epi_vsetvl(N-j,__epi_e32,__epi_m1);
			#pragma clang loop unroll_count(32)
			for (uint32_t k = 0; k < K; ++k) {

				auto A_i_k = ...;
				auto B_k_j = bfloat16_unpack(&B[k * N + j],gvl);

				accum = __builtin_epi_vfmacc_4xf32(accum, A_i_k, B_k_j,gvl);
			}

			bfloat16_pack(&C[i * N + j], accum,gvl);
	}
}
\end{lstlisting}

\begin{lstlisting}[float=*t, label=lst:unpack,caption={Vectorized bfloat16 $\rightarrow$ binary32 load}]

inline __epi_4xf32 bfloat16_unpack(const signed short int *data,
                            unsigned long grantedLength) {
  // Expand 16-bit integer StorageType to 32-bit integer
  // 1. Load vector
  __epi_4xi16 data_v = __builtin_epi_vload_4xi16(data, grantedLength);

  // 2. Expand using widening
  __epi_4xi32 wdata_v = __builtin_epi_vwadd_4xi32(
      data_v, __builtin_epi_vmv_v_x_4xi16(0, grantedLength), grantedLength);

  // Conversion operations
  __epi_4xi32 res_v = __builtin_epi_vsll_4xi32(
      wdata_v, __builtin_epi_vmv_v_x_4xi32(16, grantedLength), grantedLength);

  // Reinterpret 32-bit integers as vector wide type
  return (__epi_4xf32)res_v;
}
\end{lstlisting}

\begin{lstlisting}[float=*t, label=lst:pack,caption={Vectorized binary32 $\rightarrow$ bfloat16 store }]
inline void bfloat16_pack(signed short int *dst, __epi_4xf32 data, 
                   unsigned long grantedLength) {
  // Reinterpret 32-bit floats into 32-bit integers
  __epi_4xi32 idata_v = (__epi_4xi32)data;

  // Conversion operations + narrowing operation
  __epi_4xi16 res_v = __builtin_epi_vnsrl_4xi16(
      idata_v, __builtin_epi_vmv_v_x_4xi16(16, grantedLength), grantedLength);

  // Store back the value in memory
  __builtin_epi_vstore_4xi16(dst, res_v, grantedLength);
}
\end{lstlisting}

\subsection{Experimental results}

We tested the approach described in the previous sections to the GEMM algorithm. In detail, we run the binaries inside the Vehave simulator, running on a RISC-V virtual machine. Then, we captured the trace output of our binaries running on the Vehave simulator and we fed them to the MUSA simulator. We used $128\times128, 256\times256, 512\times512$ sizes for the matrices in the experiments. Furthermore, we varied the length of the vector registers in $4096,8192,16384$ bits to assess the impact of this parameter on the proposed approach. We firstly measured the number of committed vector instructions and then, exploiting a RISC-V processor timing model we obtained the average latency for each committed instruction, so that we could assess the latency impact of our approach in the different configurations.

\begin{figure*}
    \centering
    \includegraphics[width=\linewidth]{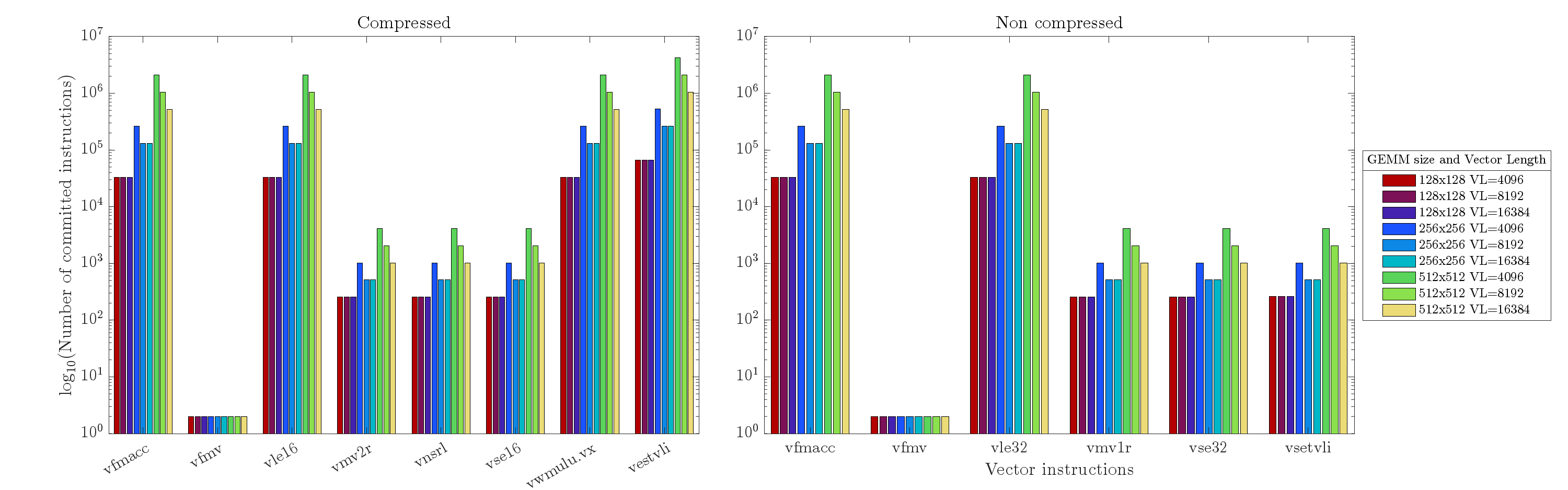}
    \caption{Number and type of vector instructions for a GEMM execution with the two approaches.}
    \label{fig:gemm_instr}
\end{figure*}

Figure \ref{fig:gemm_instr} reports the number of committed instructions required to complete the different algorithm, both with compressed and un-compressed approach. We used a logarithmic scale for the y-axis to fit both frequent and less frequent instructions inside the same plot. From the comparison we can see that the compressed approach has additional instructions related to the compression and decompression phases. In particular, the new instructions are:
\begin{itemize}
    \item \texttt{vnsrl}: logical shift right with narrowing from 32-bit to 16-bit. This instruction is used to directly compress a 32-bit binary32 vector into a 16-bit bfloat16 one, discarding the least 16 significant bits of the fractional part.
    \item \texttt{vwmul}: this instruction has the opposite effect of the previous one: this is used to directly decompress a 16-bit bfloat16 vector into a 32-bit binary32 one, by padding 16 zeroes to the right of the bfloat number (actually multiplying by $2^{16}$). 
    \item \texttt{vsetvli}: this instruction is automatically inserted by the compiler whenever we change from handling 32-bit elements to 16-bit ones and vice-versa. This instruction is completely unrelated to the proposed approach and it is specific of the vector architecture being used.
\end{itemize}
Another difference that emerges from the comparison is that the compressed approach has the same amount of load and stores but carried out on 16-bit elements, whereas the uncompressed approach resorts to load and store uncompressed elements on 32-bits. 

Since the number of instructions cannot solely measure the impact of this difference we also simulated the program using the Tasksim RISC-V simulator. For this task we only considered the largest workload (i.e. $512 \times 512$ GEMM) with $16Kbits$ vector registers. The simulator embeds a timing model of a RISC-V processor and uses the trace files output by Vehave to produce detailed  statistics and in particular, the overall number of computing cycles elapsed to complete program execution. This metric is used to assess the improvement of the compressed approach ($\text{cycles}_c$) over the non-compressed one ($\text{cycles}_u$). We define the improvement metric as follows.
\begin{equation}
    \text{improvement} = 100 \times \left ( 1 - \frac{\text{cycles}_c}{\text{cycles}_u} \right )
\end{equation}

The improvement metric is impacted by a series of parameter of the simulator:
\begin{itemize}
    \item L1 data cache size and latency
    \item Memory bandwidth and latency
    \item Latency of the overhead instructions used for compression and decompression (i.e. \texttt{vnsrl}, \texttt{vwmulu}).
\end{itemize}

\begin{figure}
    \centering
\includegraphics[width=\linewidth]{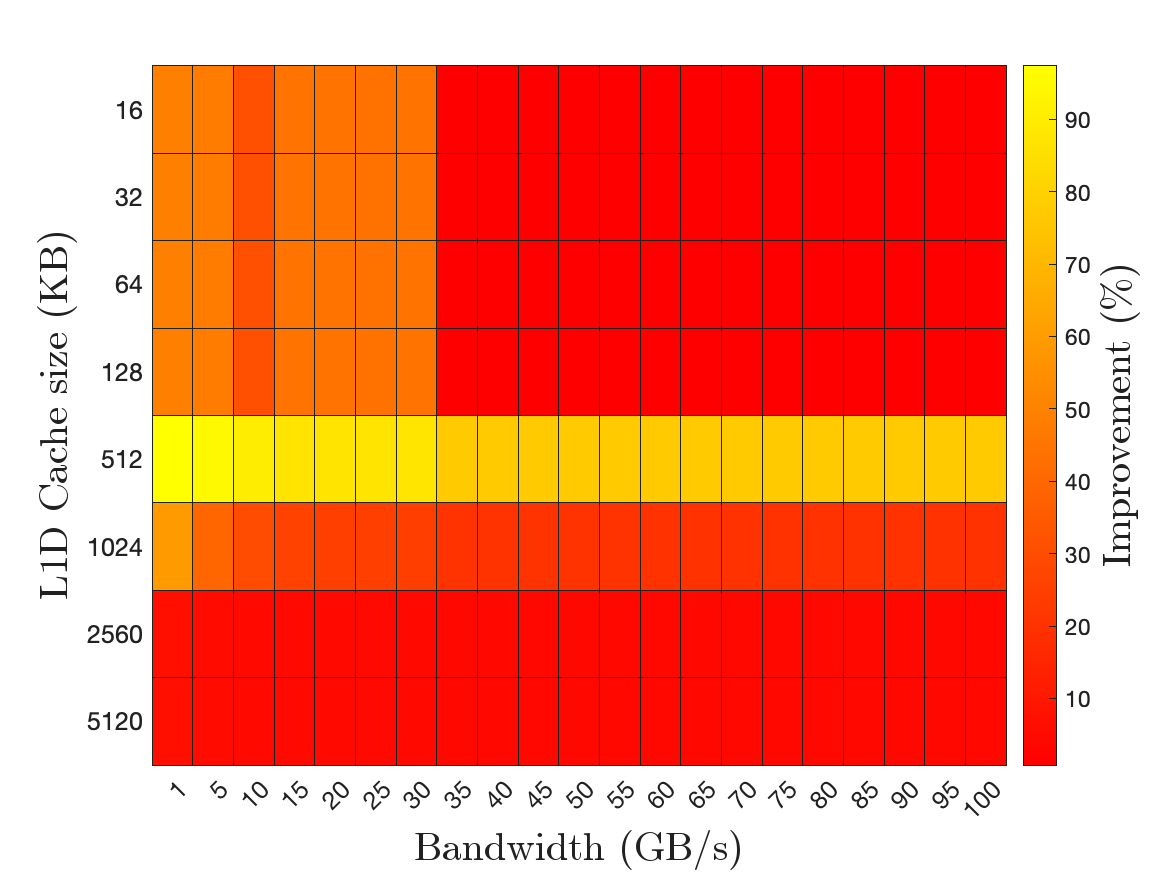}
    \caption{Improvement using the compression approach when varying the memory bandwidth and L1 cache size.}
    \label{fig:impr_vs_bw_l1size}
\end{figure}

Firstly, we considered L1 cache size and memory bandwidth fixing memory latency at $60ns$ and cache latency at $20ns$. Figure \ref{fig:impr_vs_bw_l1size} shows the variation of the improvement when using the compression approach when changing the memory bandwidth and cache size. As reported, the peak improvement is associated with a cache size of $512Kb$. In particular, with this cache size, the improvement spans from $77\%$ with a $100GB/s$ bandwidth up to $97\%$ with a $1GB/s$ bandwidth. Table \ref{tab:impr_vs_bw_cs} reports the details of the improvement with selected cache and bandwidth configurations.

\begin{table}[]
\caption{Improvement of the compression approach with low-end and high-end memory bandwidth and cache sizes.}
\label{tab:impr_vs_bw_cs}
\centering
\begin{tabular}{l|ll} 
\cline{2-3}
           & \multicolumn{2}{l}{Memory bandwidth}  \\ 
\cline{2-3}
Cache size & 1GB/s  & 100GB/s                      \\ 
\hline
512KB      & 97.3\% & 77.3\%                       \\
1MB        & 59.3\% & 20.4\%                       \\
2.5MB      & 6.8\%  & 4.1\%                        \\
\hline
\end{tabular}
\end{table}

\begin{figure*}[t]
    \centering
    \includegraphics[width=\linewidth]{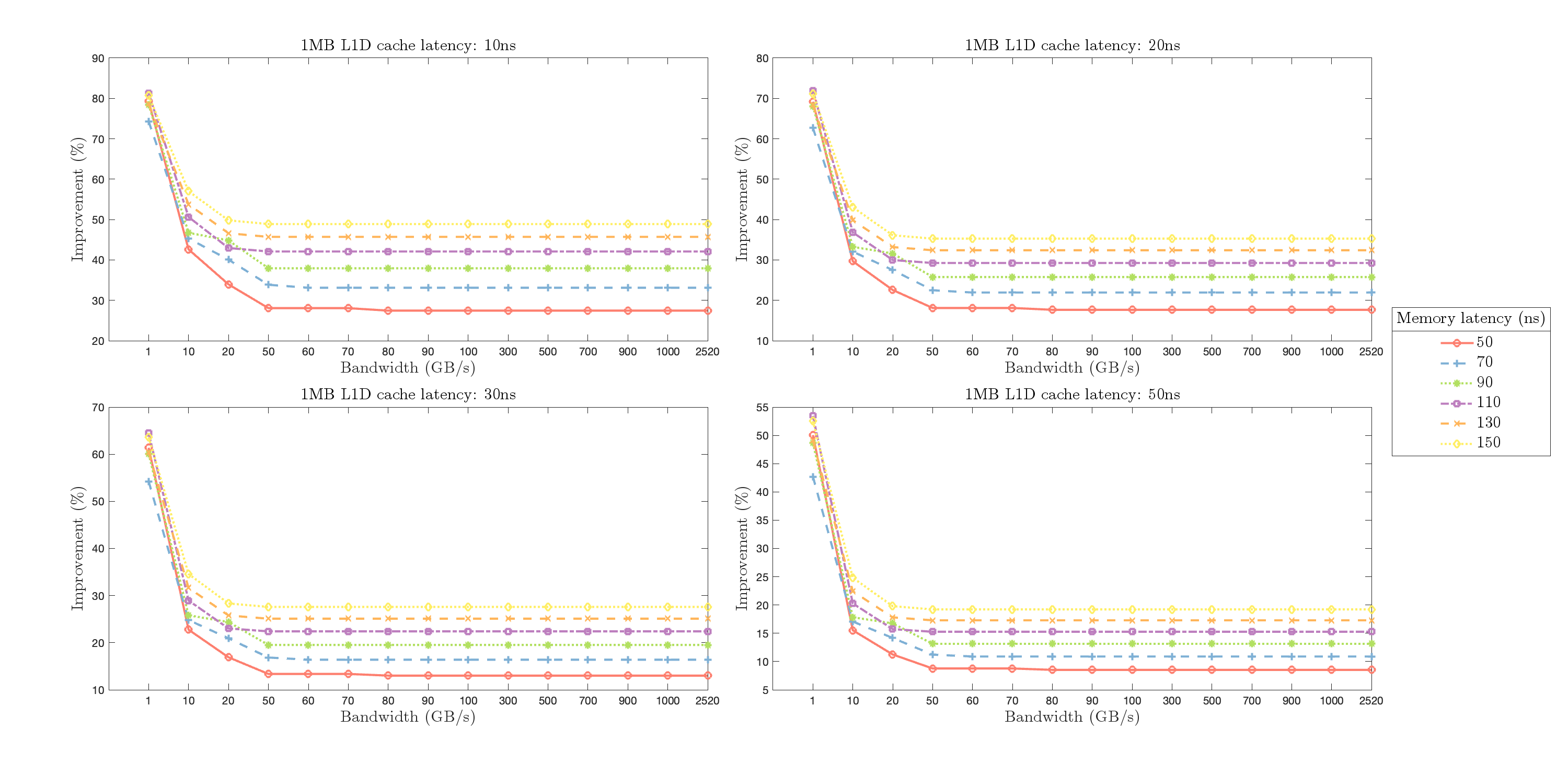}
    \caption{Improvement using the compressed approach compared to the non-compressed one when varying the memory bandwidth and L1 cache latency and memory latency, with a fixed 1MB cache size.}
    \label{fig:impr_vs_lat_bw}
\end{figure*}

Secondly, we fixed a reasonable cache size of $1MB$ and then varied the L1 cache latency, memory latency and bandwidth to analyse their impact. Figure \ref{fig:impr_vs_lat_bw} shows the variation of the improvement between the compressed and un-compressed approach. As expected, the lower the bandwidth the higher the benefits of compressing data to be transferred. Moreover, the higher the memory latency, the higher the impact of the compression in terms of improvement. Furthermore, the workload considered does not benefit from increasing the bandwidth over $50GB/s$ with these configurations.

Finally, we considered the latency of the instructions executed for the compression and decompression. We selected beforehand the following configurations for the architecture:
\begin{itemize}
    \item Memory bandwidth: $[10,50,100] GB/s$
    \item Cache latency: $15ns$
    \item Memory latency: $[50,70,90,110,130,150]ns$
    \item Cache sizes: $[512KB,1MB,2.5MB]$
    \item Instruction latency: $[1,10,20,30,40,50]$ cycles.
\end{itemize}
Figure \ref{fig:impr_vs_ins_lat} shows the impact of the different architecture parameters on the improvement when using the compression approach. As expected and theorised in the previous section, the latency of compression and decompression instruction has a great impact on the improvement metric. In particular, an instruction latency over $20ns$ corresponds to a sharp drop in the improvement, even nullifying it for bigger caches and bandwidths (as we can see in the third column of Figure \ref{fig:impr_vs_ins_lat} . Analysing this behaviour is fundamental to understand the requirements of a compressed real number representation to be used with this approach. Indeed, the seamless conversion between bfloat16 and fp32 can be obtained with the cost of a single ALU operation. Concluding, the information gathered and represented in Figure \ref{fig:impr_vs_ins_lat} can serve as a comparison with other compressed real number formats such as posits in terms of the achievable improvement when used to compress memory data transfers.

\subsection{The impact of posit de/compression} 
\begin{figure*}[t]
    \centering
    \includegraphics[width=\linewidth]{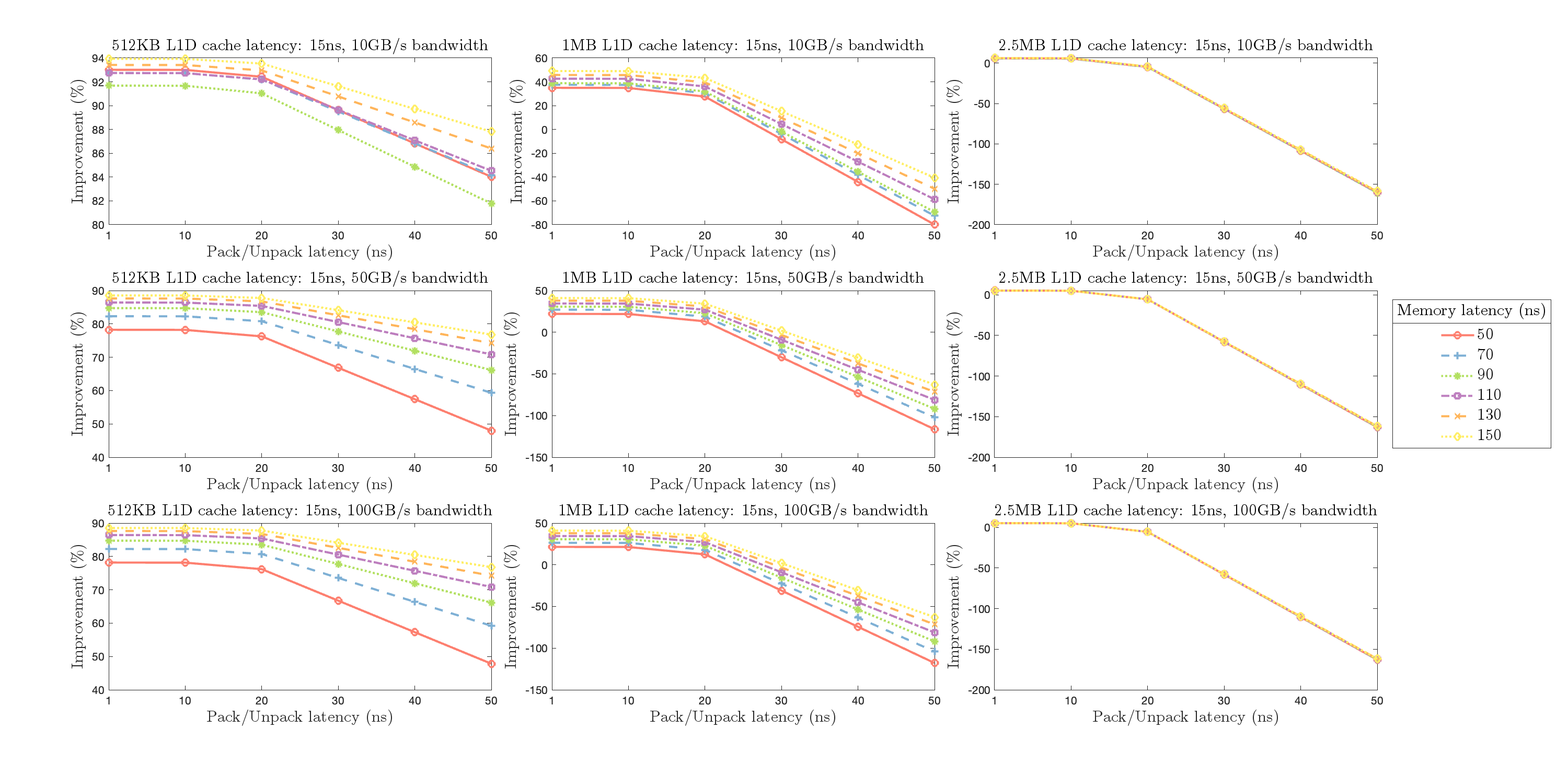}
    \caption{Improvement using the compressed approach when varying architectural parameters and compression/decompression instruction latency.}
    \label{fig:impr_vs_ins_lat}
\end{figure*}

Without modifying the existing instruction set architecture, the conversion between a 16-bit posit and a 32-bit floating point has a much higher cost in terms of number of instruction and therefore in terms of overall latency. However, if we managed to have a single instruction that was able to perform this operation (with a dedicated hardware unit) with a latency that is compatible with the ones shown in Figure \ref{fig:impr_vs_ins_lat} we could employ also 8 or 16-bit posits for this task.

\section{Conclusions}\label{sec:concl}

In this paper, we have demonstrated the implementation of a GEMM kernel using compressed real number formats on a RISC-V vector architecture. The significance of this kernel lies in its potential to serve as a demonstrator for other neural network workloads. Additionally, we have presented a comprehensive stack of tools designed for the development and analysis of RISC-V vectorized applications.

Our investigation included an analysis of the behavior of the bfloat16 format when utilized as a compressed format for storing image and weight data. Leveraging the seamless conversion between bfloat16 and binary32, we achieved a substantial speedup of approximately $90\%$ compared to an uncompressed solution that uses binary32 for both storage and computations.

We believe our contribution is twofold: firstly, we have demonstrated another promising use case of reduced precision data types, specifically bfloat16; secondly, by adopting a co-design approach with simulation tools, we were able to provide valuable feedback to RTL designers, ultimately leading to improvements in the architecture and enabling the efficient utilization of compressed data type formats.

Our work exemplifies the effectiveness of the co-design approach, incorporating simulations, in obtaining early feedback crucial for the hardware design stage. This approach proves to be highly useful for optimizing and refining designs, laying the groundwork for future advancements in this domain.

\bibliographystyle{IEEEtran}
\bibliography{j7}

\EOD

\end{document}